\documentclass[sigconf]{acmart}
\usepackage{xcolor}
\usepackage{caption} %Added by CS
\usepackage{subcaption} %Added by CS
\renewcommand\footnotetextcopyrightpermission[1]{} % removes footnote with conference information in first column
\renewcommand\acmConference[1]{}

\begin{document}

\title{In Situ Framework for Coupling Simulation and Machine Learning with Application to CFD}
% In Situ Framework for Coupling Simulation and Machine Learning with Application to CFD Domain
% A Scalable Infrastructure for In Situ Coupling of Simulations and Machine Learning
% In Situ Coupling of Simulations and Machine Learning: Scaling Benchmarks and Applications to CFD

\author{Riccardo Balin}
\email{rbalin@anl.gov}
\affiliation{%
  \institution{Argonne National Laboratory}
  \city{Lemont}
  \state{IL}
  \country{USA}
}

\author{Filippo Simini}
\email{fsimini@anl.gov}
\affiliation{%
  \institution{Argonne National Laboratory}
  \city{Lemont}
  \state{IL}
  \country{USA}
}

\author{Cooper Simpson}
\email{cooper.simpson@colorado.edu}
\affiliation{%
  \institution{University of Colorado Boulder}
  \city{Boulder}
  \state{CO}
  \country{USA}
}

\author{Andrew Shao}
\email{andrew.shao@hpe.com}
\affiliation{%
  \institution{Hewlett Packard Enterprise}
  \city{Seattle}
  \state{WA}
  \country{USA}
}

\author{Alessandro Rigazzi}
\email{alessandro.rigazzi@hpe.com}
\affiliation{%
  \institution{Hewlett Packard Enterprise}
  \country{Switzerland}
}

\author{Matthew Ellis}
\email{matthew.ellis@hpe.com}
\affiliation{%
  \institution{Hewlett Packard Enterprise}
  \city{Seattle}
  \state{WA}
  \country{USA}
}

\author{Stephen Becker}
\email{stephen.becker@colorado.edu}
\affiliation{%
  \institution{University of Colorado Boulder}
  \city{Boulder}
  \state{CO}
  \country{USA}
}

\author{Alireza Doostan}
\email{alireza.doostan@colorado.edu}
\affiliation{%
  \institution{University of Colorado Boulder}
  \city{Boulder}
  \state{CO}
  \country{USA}
}

\author{John A. Evans}
\email{john.a.evans@colorado.edu}
\affiliation{%
  \institution{University of Colorado Boulder}
  \city{Boulder}
  \state{CO}
  \country{USA}
}

\author{Kenneth E. Jansen}
\email{kenneth.jansen@colorado.edu}
\affiliation{%
  \institution{University of Colorado Boulder}
  \city{Boulder}
  \state{CO}
  \country{USA}
}

\renewcommand{\shortauthors}{Balin, et al.}

\begin{abstract}
%Recent years have seen many successful applications of machine learning (ML) to facilitate computations of complex turbulent flows. As simulations grow in complexity and size, generating new databases of training data for traditional offline learning creates an I/O and storage bottlenecks. Additionally, performing inference with ML models at runtime requires non-trivial coupling of ML framework libraries with the simulation code. This work offers a solution to both limitations by simplifying this coupling and enabling in situ training and inference workloads on heterogeneous high performance computing clusters. Leveraging the SmartSim open-source tool, the presented framework deploys a database to store training and inference data, useful metadata, and ML models in memory, thus circumventing the file system entirely. On the Argonne Polaris supercomputer, we demonstrate near-perfect scaling efficiency up to the full size of the machine of the data transfer and ML inference costs thanks to a novel deployment of the AI-aware database which is co-located on the same nodes as the simulation. Moreover, we apply our framework to train an autoencoder model from a turbulent flow simulation, showing that the overhead incurred due to the data transfer is negligible relative to a solver time step and training epoch.
%
Recent years have seen many successful applications of machine learning (ML) to facilitate fluid dynamic computations. As simulations grow, generating new training datasets for traditional offline learning creates I/O and storage bottlenecks. Additionally, performing inference at runtime requires non-trivial coupling of ML framework libraries with simulation codes. This work offers a solution to both limitations by simplifying this coupling and enabling in situ training and inference workflows on heterogeneous clusters. Leveraging SmartSim, the presented framework deploys a database to store data and ML models in memory, thus circumventing the file system. On the Polaris supercomputer, we demonstrate perfect scaling efficiency to the full machine size of the data transfer and inference costs thanks to a novel co-located deployment of the database. Moreover, we train an autoencoder in situ from a turbulent flow simulation, showing that the framework overhead is negligible relative to a solver time step and training epoch. 
\end{abstract}

\maketitle
%\printacmref

\section{Introduction}
\label{sec:Intro}

\begin{figure}
    \begin{subfigure}{\columnwidth}
        \centering
        \includegraphics[width=\columnwidth]{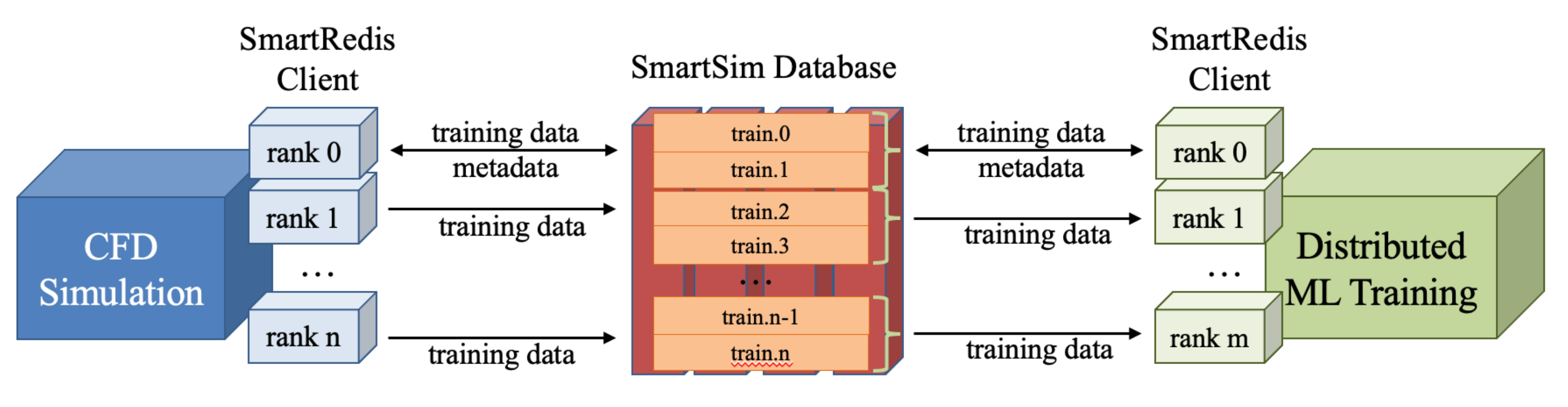}
        \caption{In situ training.}
        \label{fig:framework_train}
    \end{subfigure}
    \vfill
    \begin{subfigure}{\columnwidth}
        \centering
        \includegraphics[width=\columnwidth]{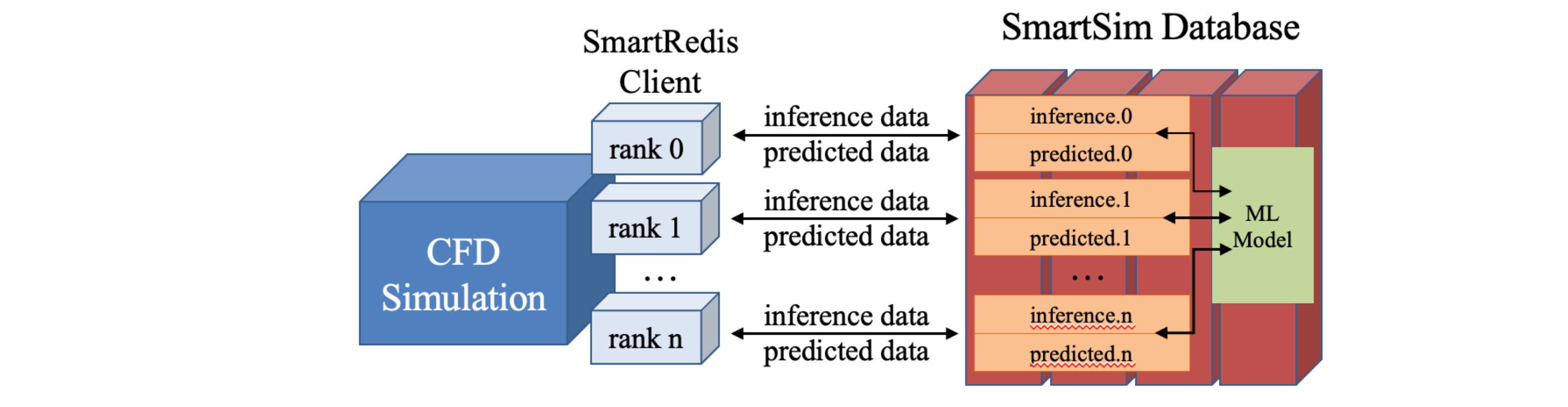}
        \caption{In situ inference.}
        \label{fig:framework_inf}
    \end{subfigure}
    \caption{Framework for in situ training and inference coupling CFD simulation codes to ML workloads with a SmartSim database and SmartRedis client library.}
    \label{fig:framework}
\end{figure}

The last decade has seen a number of successful demonstrations of the use of machine learning (ML) to facilitate computational fluid dynamic (CFD) simulations \cite{Brunton2020}. 
These works span a wide range of applications, including the development of turbulent closure models of various forms and fidelities \cite{Peters2022,Prakash2022,Stoffer2021,Bae2022},  compression of flow states \cite{Glaws2020,Doherty2023}, turbulent inflow generation \cite{Yousif2022}, flow control \cite{Li2022}, and even surrogate modeling of the Navier-Stokes equations \cite{Pfaff2021,Stronisch2023}.
In all these cases, the ML models were trained using the traditional offline (or \textit{post hoc}) approach, wherein the training data is produced beforehand by a high-fidelity simulation, usually by a different research group \cite{Bae2022,Prakash2022}, and stored in private or public databases \cite{JHTDB-URL,NTMR-URL}.

While more straightforward, the offline data pipeline imposes some important restrictions to the development and deployment of ML models which can generalize to new geometries and physical regimes. 
First, generating, storing and maintaining the training database is a costly exercise since it involves storing many high-fidelity solution snapshots computed with direct numerical simulations (DNS) \cite{JHTDB-URL,NTMR-URL}. 
Second, model training is limited to the few datasets available to the research community in terms of variety of the flow physics (which mostly covers canonical problems lacking the complexity of the target flows), variety of the flow variables saved from the simulation, and quality of the data. Together, these two factors prevent models from being fine-tuned on the specific problem classes of interest and lead to deployment on out-of-sample data and reduced accuracy.
Lastly, the offline training approach does not directly address the non-trivial task of coupling the simulation code with ML framework libraries in order to perform inference at runtime. This final hurdle in the workflow often precludes ML models from being deployed in production runs and results in bespoke implementations within a particular CFD code.  

To solve the limitations imposed by the offline training pipeline, in situ (or online) ML approaches have recently emerged as an attractive solution for both training and inference workloads. 
In the case of in situ training, the model learning and simulation are run together, storing the training data in-memory as it is being produced \cite{Maulik2020,Maulik2022,Sirignano2023,Madireddy2021,Kurz2023}. Therefore, by circumventing the file system and training on the desired flow problem, in situ methods eliminate the need for expensive databases and reduce the differences between distributions of the training and prediction datasets.
Similarly, during in situ inference, the ML model is loaded in-memory and evaluated by the simulation as it produces new inference data on which to make predictions \cite{Maulik2020,Ott2020,Brewer2021}. 

Following the terminology established by the in situ visualization and analysis literature \cite{Kress1025,Childs2020,Rodero2016,Ramesh2022}, in situ ML approaches can be grouped into tightly-coupled and loosely-coupled implementations. 
In the former, also referred to as \textit{in line}, the simulation and ML components time-share the same computational resources or processes, thus alternating between advancing the solution and running inference or training.
\citet{Sirignano2023} and \citet{Maulik2022} used tightly-coupled approaches for in situ training of turbulence closure models and to perform singular value decomposition (SVD) of solution states, respectively. For inference, in line approaches embed Python or an ML framework library within a C++ or Fortran simulation code \cite{Maulik2020,Geneva2019,Ott2020}. 
In the loosely-coupled implementation, also referred to as \textit{in transit}, the simulation and ML components run concurrently on separate dedicated compute resources and communicate data between each other asynchronously. \citet{Kurz2023} performed in transit training of a closure model with deep reinforcement learning.
Hybrid implementations have also been adopted for in situ inference, where distinct resources are used for the simulation and ML component however the model evaluation is not run asynchronously and instead blocks the progress of the simulation \cite{Brewer2021,Partee2022}.

In this paper, we propose a framework that facilitates the coupling of CFD simulation codes with ML by enabling in situ training and inference capabilities. 
In particular, we leverage the open source tool called SmartSim \cite{SmartSim-URL} to deploy a database which can store training and inference data, useful metadata, and ML models in-memory for the duration of the run. Alongside the database, which can be deployed either on-node or off-node, we make use of the SmartRedis API library \cite{SmartRedis-URL} to transfer data between the workflow components and evaluate ML models on heterogeneous high performance computing (HPC) architectures.
This combination of SmartSim and SmartRedis offers four key advantages. First, the database allows in situ training to be loosely-coupled and fully asynchronous, leading to negligible idle time in the simulation and ML training applications. Second, our novel approach launches a database co-located on the same nodes as the simulation and ML training, thus achieving perfect scaling efficiency. 
Third, the SmartRedis API facilitates the integration of any existing ML training workload into the in situ infrastructure by simply modifying the data loader. 
Lastly, the SmartRedis API allows the simulation to remain agnostic of the model structure and framework library used for training while performing in situ inference, thus simplifying coupling the two components, and enables inference on GPU hardware from CPU-only simulation codes.

Moreover, we apply our framework to perform in situ training of an autoencoder for compression of solution states.  
The PHASTA flow solver \cite{WhiJan01} is integrated with the SmartRedis library and used to generate data by DNS of a turbulent flat plate boundary layer on a 36 million element grid. Concurrently, an autoencoder model adapted from \citet{Doherty2023} is trained on the live simulation data in parallel on 160 GPU, producing the encoder and decoder fine-tuned on this flow problem.
Inference with the encoder can then performed by PHASTA at runtime on future time steps, thus enabling a much richer time history of the solution field to be saved to disk.
%Inference with the encoder is then performed by PHASTA at runtime on future time steps, thus enabling a much richer time history of the solution field to be saved to disk. 

The remainder of this paper makes the following contributions:
\begin{itemize}
    \item The infrastructure for coupling simulation and ML as outlined above is described in detail, offering the novel deployment of an on-node or co-located database. This contribution distinguishes this work from the previous uses of SmartSim and SmartRedis \cite{Partee2022,Kurz2023,Brewer2021}.
    \item Through a series of tests, the perfect scaling efficiency of the data transfer and model evaluation overheads associated with in situ training and inference is demonstrated up to full size of the Polaris supercomputer.
    \item The in situ infrastructure is showcased on the realistic application of training an autoencoder for compression of turbulent flow states.
\end{itemize}

The paper is organized as follows. Section~\ref{sec:Infra} describes in detail the in situ framework designed to facilitate in situ training and inference from simulation codes. Section~\ref{sec:Scaling} covers the performance and scaling tests carried out on the Argonne Polaris supercomputer. The framework is then applied to train the autoencoder with solution data from a turbulent flow simulation in Section~\ref{sec:Phasta}. Finally, Section~\ref{sec:Conclusion} offers some concluding remarks and directions for future research.

% Notes:
% 1. The title key words are: in situ, coupling CFD simulation and ML. So there should be a lit review and motivation for both.
% 2. CFD + ML: I would start here, as the abstract does. Start with past work of CFD+ML, which did traditional offline training and some couplings for inference. Motivate the need for in situ, which is in fact the motivation for this work.
% 3. In situ: There have been applications of in situ stuff and coupling of simulation and ML, some more general and related to viz some more specific to ML and CFD. The goal here is to put our work in the context of other in situ work, use the right terminology.
% 4. SmartSim: Introduce SmartSim and discuss other papers that used this tool.
% 4. Bullets with 3 main contributions that this paper makes

%require multi-terabyte databases \cite{JHTDB-URL}, which are expensive to fund, mostly contain solutions to canonical flow problems, and do not allow streaming access to the data. 

% Examples of ML for CFD
% 
\section{In Situ Framework}
\label{sec:Infra}

\subsection{Framework Description}

%\begin{figure}
%    \begin{subfigure}{\columnwidth}
%        \centering
%        \includegraphics[width=\columnwidth]{figures/diagrams/framework_training_diagram.pdf}
%        \caption{In situ training.}
%        \label{fig:framework_train}
%    \end{subfigure}
%    \vfill
%    \begin{subfigure}{\columnwidth}
%        \centering
%        \includegraphics[width=\columnwidth]{figures/diagrams/framework_inference_diagram.pdf}
%        \caption{In situ inference.}
%        \label{fig:framework_inf}
%    \end{subfigure}
%    \caption{Framework for in situ training and inference coupling CFD simulation codes to ML workloads with a SmartSim database and SmartRedis client library.}
%    \label{fig:framework}
%\end{figure}

The framework proposed in this work to enable in situ learning and inference of ML models from traditional simulation codes  consists of four major components -- a data producer, a data consumer, a database and a communication library. A schematic of these components is shown in Figure ~\ref{fig:framework}.
As the terms suggest, the data producer, in this case a CFD simulation code, generates the training and inference data, while the data consumer, a distributed training workload or inference module, operates on that data to learn or obtain predictions from an ML model.
The third and key component of the framework is an in-memory database, which stores any information shared by the producer and consumer for the duration of the run, thus avoiding the need to write to disk. 
Lastly, a communication library provides the API to interact with the database and the data or models stored within it from the simulation and training applications. 

The database and communication library are provided by SmartSim \cite{SmartSim-URL}, which is an open source tool developed by Hewlett Packard Enterprise specifically designed to provide scientific codes access to ML facilities at runtime.
There are two core components to SmartSim -- the infrastructure library (IL) and a client communication library called SmartRedis \cite{SmartRedis-URL}. The main role of the IL is to provide API to start, monitor and stop HPC jobs from Python and to deploy the in-memory database. The database is a key-value store with a shared-nothing architecture enabling low-latency access to many clients in parallel. Both Redis \cite{Redis-URL} and KeyDB \cite{KeyDB-URL} databases can be used with the SmartSim API. 
SmartRedis, on the other hand, provides clients that can connect to the database from Fortran, C, C++, and Python, thus serving most CFD simulation codes and ML workloads. The client API provides send/retrieve semantics for data communication, which occurs over the TCP/IP stack, as well as an RPC-like API for invoking JIT-traced Python and ML runtimes, such as running models for inference.

During in situ training, as outlined in Figure~\ref{fig:framework_train}, all four components of the framework are distinct from each other. The training data flows from the simulation, where it is generated, through the database to the ML workload, where it is consumed for training. The simulation and training applications interact independently and asynchronously only with the database making use of the SmartRedis API, but not directly with each other. 
The database, therefore, on top of storing the training dataset for the duration of the job, ensures a loose-coupling between the two components. As such, it benefits from the many advantages of loosely-coupled and \textit{in transit} approaches \cite{Kress1025,Rodero2016,Ramesh2022}, most notably the fact that the simulation is allowed to run effectively unaltered, carrying out the compute intensive PDE integration with the only blocking operation being the data transfer to the database. Similarly, the distributed training workload remains largely untouched as the data loader gathers batches from the database rather than from files stored on the disk.

In contrast, as depicted in Figure~\ref{fig:framework_inf}, during in situ inference the ML component is not distinct from the database. The framework leverages the RedisAI module \cite{RedisAI-URL} to execute ML models, both on CPU and GPU, and manage their data within the Redis or KeyDB databases. By providing API to call the RedisAI module, the SmartRedis library enables the CFD simulation to load the model, transfer the inference and predicted data to and from the database, and perform inference. Similarly to the other uses of SmartSim for in situ inference \cite{Brewer2021,Partee2022}, this can be considered a hybrid approach since the simulation is idle during the data data transfer and model evaluation, however the model is run on separate compute resources.

It is important to note that while this paper focuses on in situ ML for CFD applications, our framework is not limited to this scientific domain. It was in fact designed to be applicable to any field of computational science and engineering requiring such a coupling. Furthermore, while the description above mentions a single data producer and consumer, this is by no means a requirement of our framework. Multiple simulations and ML, data analysis and visualization workloads can be launched concurrently while communicating asynchronously with the database, allowing for creative workflows tailored to the specific needs of various applications.

\begin{figure}
    \begin{subfigure}{\columnwidth}
        \centering
        \includegraphics[width=\columnwidth]{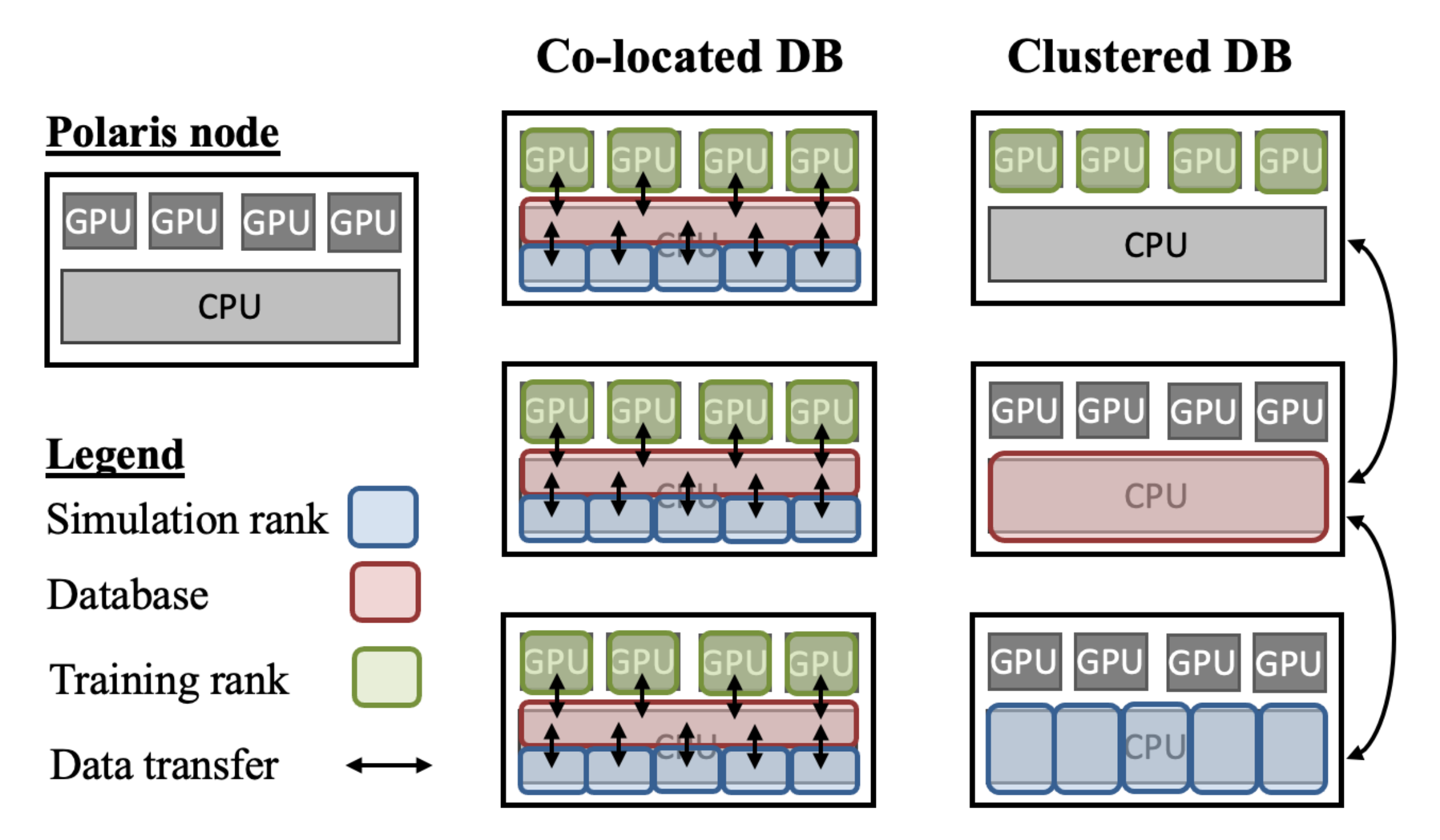}
        \caption{In situ training.}
        \label{fig:deployment_train}
    \end{subfigure}
    \vfill
    \begin{subfigure}{\columnwidth}
        \centering
        \includegraphics[width=\columnwidth]{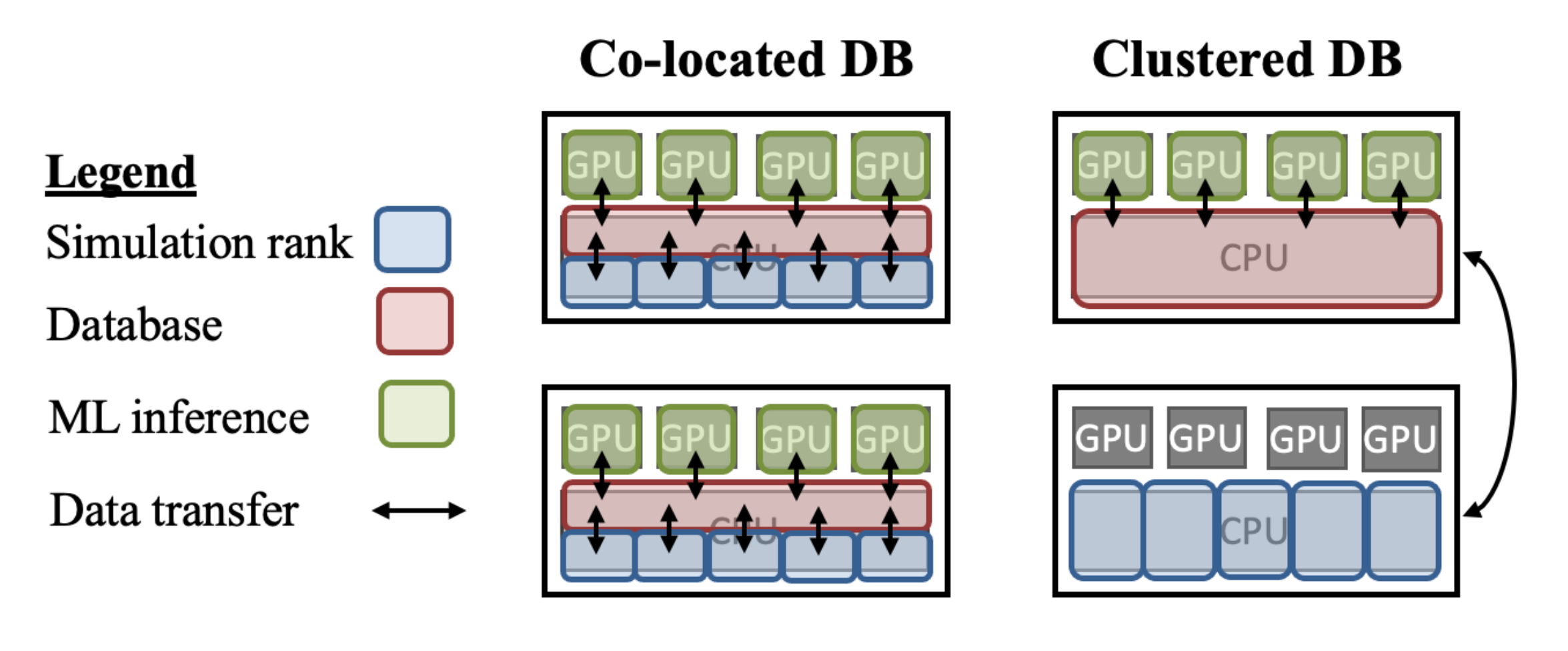}
        \caption{In situ inference.}
        \label{fig:deployment_inf}
    \end{subfigure}
    \caption{Deployment of the framework for in situ training and inference on the Polaris nodes with a co-located and clustered database (DB).}
    \label{fig:deployment}
\end{figure}

\subsection{Implementation}

Implementing the in situ framework requires two key steps. First, a driver program is built in order to manage and launch the various components of the workflow. This is a Python script which makes use of the SmartSim IL API to set up and run the database, CFD simulation and, in case of training, the distributed ML workload. The IL API interfaces with the job scheduler and the parallel launcher on the local machine or supercomputing cluster in order to execute these tasks. The driver is also responsible for the specific deployment strategy used to launch the database, as described in Section~\ref{sec:deployment}.

% Maybe add some code snippets if there is time/space, although the code for this work will be available.

The second step is integrating the SmartRedis API into the CFD solver and distributed training codes to enable the coupling of the two. As shown in Figure~\ref{fig:framework_train}, a SmartRedis client is initialized by each rank of the simulation code, effectively establishing the connection to the database. Then, the ranks send their contribution of the training data (assuming the use of domain decomposition) to the database with a unique key, resulting in $n+1$ distinct tensors being stored in memory. The tensor key can be unique to both the rank and the time step number, thus avoiding the overwriting of previously generated data. 
Similarly, the distributed training application initializes a SmartRedis client on each of its ranks and gathers the data before each epoch by simply modifying the existing dataloaders. It is often the case that fewer ranks are used for the training application than the simulation, resulting in the $m+1$ ML ranks retrieving multiple tensors from the database at random. 
It is important to note that the client initialization, data send, and data retrieve operations are performed with a single call to the SmartRedis API, each requiring a single line of code. This is an attractive feature of our framework as it simplifies the data communication required to couple simulations and ML and keeps the implementation effort to a minimum.

Performing in situ inference requires a few additional calls to the SmartRedis API from the CFD solver. As shown in Figure~\ref{fig:framework_inf}, the ML model is also loaded to the database and it is evaluated on the inference data to generate predictions. 
Both the SmartSim IL and SmartRedis library offer API to load an ML model from file, meaning this operation can be performed by the driver or the simulation code. Models in PyTorch, TensorFlow or ONNX format are compatible and can be offloaded to multiple GPU.
Then, inference is performed in three distinct steps: 1) each rank sends its inference data to the database with a unique key, 2) each rank evaluates the model leveraging RedisAI by specifying the unique key for the inference data, the predicted data and the device on which to run the model, and 3) each rank retrieves the predicted data. 
Once again, using the SmartRedis API a single line of code is sufficient for each step.

%Another key advantage of using a loosely-coupled approach is that it requires minimal changes to the simulation and ML programs. 
%In fact, sending and retrieving data to and from the database only require to add one line of code per call, as opposed to tightly-coupled approaches that require hundreds of lines of code edits and the inconvenience of dealing with libraries compatibility issues. 
%Specifically, when performing in-situ training, each of the $n + 1$ ranks of the CDF Simulation sends data to the SmartSim in-memeory database using one SmartRedis API call, as shown in Figure~\ref{fig:framework_train}. Data stored in the database is fetched by the $m + 1$ ML ranks with a single line of code and it is used to perform the training steps with no change to the ML training pipeline, other than for the definition of custom Dataset and DataLoaders.
%In the case of is-situ inference, shown in Figure~\ref{fig:framework_inf}, the trained JIT-traced PyTorch model is loaded into the database in order to perform inference on the data sent by the CFD Simulation. The model's outputs are then retrieved from the database by the appropriate simulation ranks with a single line of code. 

\subsection{Deployment}
\label{sec:deployment}

% Outline:
% \begin{itemize} 
%     \item Then move to implementation details. Start with some details of how the simulation and training use SmartRedis to move the data around. Use figures from slides. Maybe some code snippets.
%     \item Then move to deployment, which is clustered vs co-located and use of CPU/GPU.
% \end{itemize}

We deploy the in situ framework on the Polaris system installed at the Argonne Leadership Computing Facility (ALCF). 
The Polaris machine is an HPE Apollo 6500 Gen 10+ based system with 560 nodes. Each node comprises of 4 Nvidia A100 40GB GPUs and a single 2.8 GHz AMD EPYC Milan CPU with 32 physical cores (but configured to have 64 logical cores) and 512 GB of DDR4 RAM. The nodes are connected to each other through a Slingshot 10 interconnect network with dragonfly topology.

SmartSim offers two approaches for deploying the in-memory database (DB) -- a novel co-located one and the traditional clustered one. 
With the co-located approach, the simulation, the database and the ML component all share the compute resources on each node. The database is therefore co-located with the application codes and deployed on the same nodes. Schematics of this approach on the Polaris nodes are displayed in the left panels of Figures~\ref{fig:deployment_train} and~\ref{fig:deployment_inf}. As shown by the placement of the arrows, all data transfer between the components is contained within each node, thus eliminating any inter-node communication (except, of course, for that between the ranks of each application located on multiple nodes). Additionally, this deployment makes good use of the hardware since all compute resources on each node are fully utilized.

With the clustered approach, the simulation, database and ML component occupy distinct nodes, as depicted by the right panels in Figures~\ref{fig:deployment_train} and~\ref{fig:deployment_inf}. Note, however, that during inference, the database and ML component share the same nodes due to the use of the RedisAI module. 
This deployment clearly results in under-utilization of the compute resources, especially with CPU only simulation codes such as the one used in this work as the GPU are left idle on the nodes assigned to the simulation and database. Moreover, it forces the data transfer to be performed across nodes using the machine's network. 

It is worth noting that the co-located deployment suffers form a limitation that may be relevant for some applications or workflows. Since a distinct database is deployed on each node, the overall training data is partitioned across these nodes and the simulation and ML training ranks can only communicate with the on-node database directly. This, however, does not limit the ability to perform distributed training on the entire dataset and can be easily circumvented by communicating between the simulation and ML application ranks with MPI.

%The framework for in situ learning and inference is installed in a {\em conda} environment and launched using a python script that performs the following operations: first the in-memory database is started in the desired mode on a specified set of nodes, then both the simulation and the ML training script are launched using MPI on their respective (sub)set of nodes. 

%\subsection{Parameter Tuning}
%Describe the settings of the key parameters of the infrastructure (in the config files or in the driver scripts) that helped improve performance.
\section{Scaling Performance}
\label{sec:Scaling}

In this section we characterize the performance and scaling efficiency of the proposed in situ framework up to 464 nodes of the Polaris supercomputer. We cover the data transfer to and from the database first, thus representing training workflows, followed by inference with the ResNet50 model.
We investigate differences due to the two available database software, Redis and KeyDB, and the two deployment approaches, clustered and co-located.
Moreover, note that we utilize a reproducer for the CFD solver to replicate how a simulation code interacts with the database and makes use of the SmartRedis API during both training and inference workflows. This reproducer is written in Fortran in order to closely represent the PHASTA flow solver used in Section~\ref{sec:Phasta} as well as many other CFD codes written in this language. It is a parallel program that initializes a SmartRedis client on each rank and then iterates over a time step loop in which each rank sleeps for a few seconds to emulate the time spent integrating the governing equations, sends data to the database and then retrieves it. In the case of inference, the reproducer can also load an ML model to the database before the loop and evaluate it during each loop iteration.
Lastly, no model training is carried out in this section since the data retrieval performed by the Fortran reproducer is representative of the distributed ML interaction with the database, which gathers new batches before each epoch.

\begin{figure}
    \centering
    \includegraphics[width=\columnwidth]{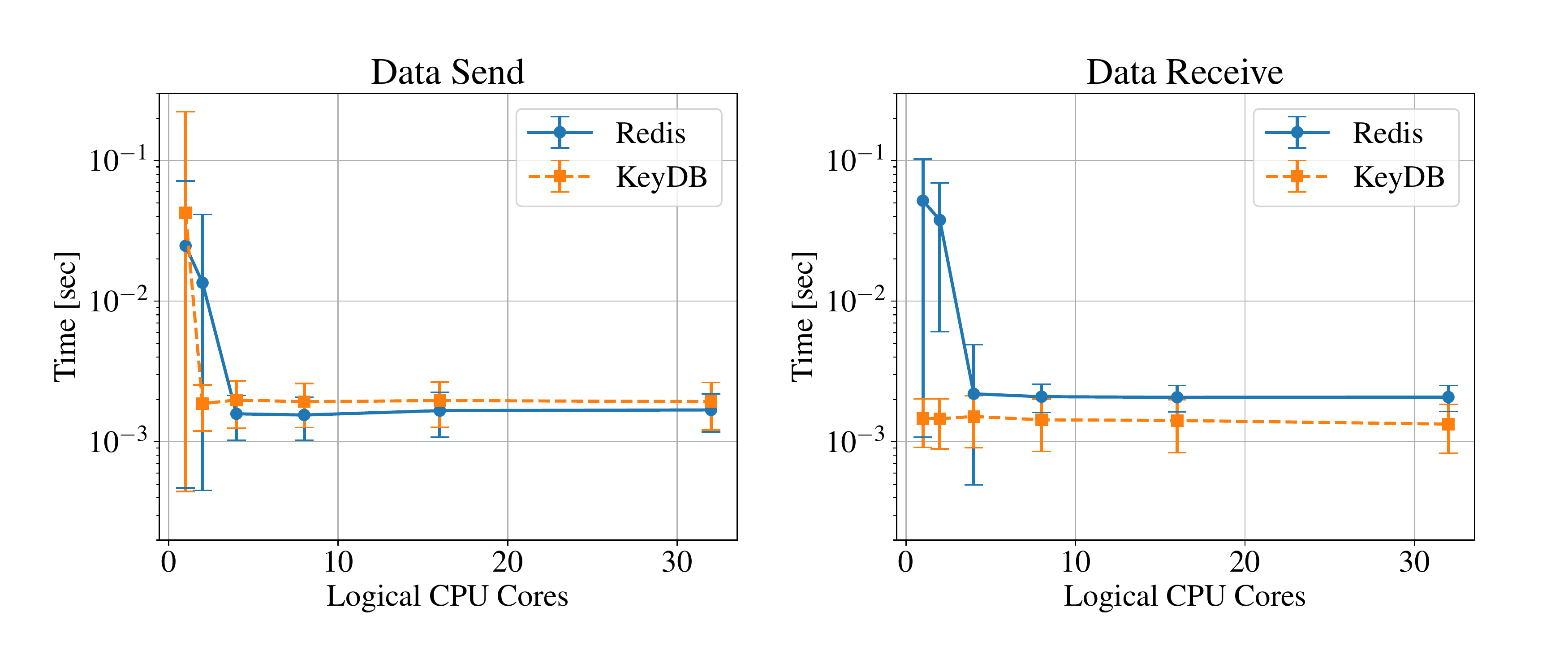}   
    \caption{Cost of data send and retrieve between simulation reproducer with increasing DB size and co-located deployment.}
    \label{fig:coDB_size}
\end{figure}

\subsection{Data Transfer Between Client and Database}
\subsubsection{Single Node Performance}
We start the performance characterization by analyzing the framework in its smallest configuration. For the co-located deployment, this means a single node is used for all components combined, however for the clustered deployment, two nodes are required for the entire framework due to the off-node placement of the database (see Figure~\ref{fig:deployment}). 

At this size, we investigate how the distribution of CPU resources between the simulation and the database impact the data transfer performance in the co-located deployment. Figure~\ref{fig:coDB_size} shows the cost of data send and retrieve operations as the number of logical CPU cores assigned to the database increases while the number assigned to the simulation reproducer is held constant at 24. 
For this performance test, each simulation rank sends and retrieves 256KB for a total of 40 iterations.
With both Redis and KeyDB, the cost of these operations is constant for DB sizes $\ge 8$ cores and the two software options achieve similar performance. KeyDB, however, is performant also with 4 logical cores. 
This result indicates that the database can occupy a small fraction of the CPU resources on each node, leaving the majority to the compute intensive CFD code and resulting in a fractional increase the number of nodes required by a simulation. 
Additionally, based on these results, all subsequent tests are performed using 8 logical cores assigned to the co-located database.

\begin{figure}
    \begin{subfigure}{\columnwidth}
        \centering
        \includegraphics[width=\columnwidth]{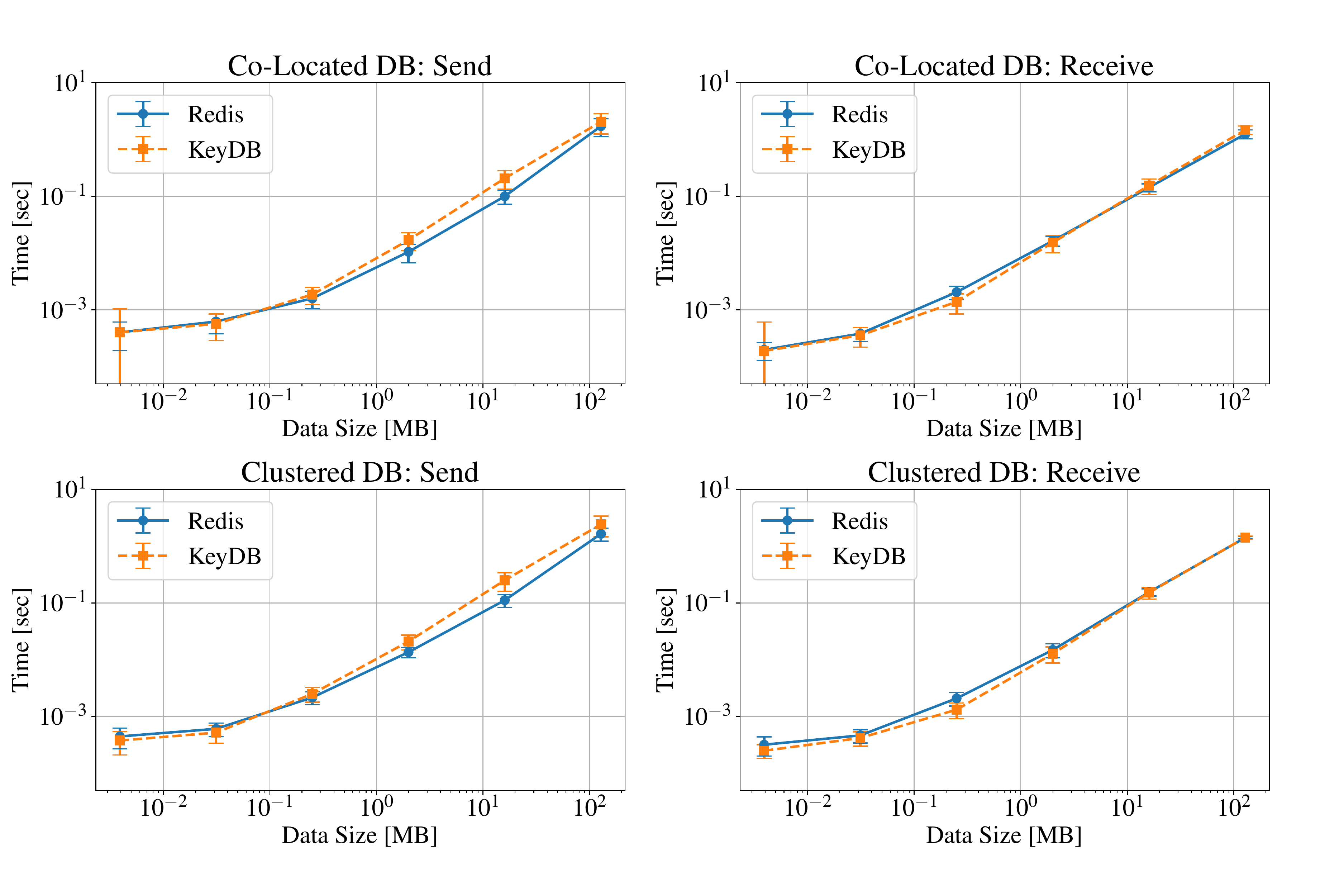}
        \caption{Time measurement.}
        \label{fig:data_size_time}
    \end{subfigure}
    \vfill
    \begin{subfigure}{\columnwidth}
        \centering
        \includegraphics[width=\columnwidth]{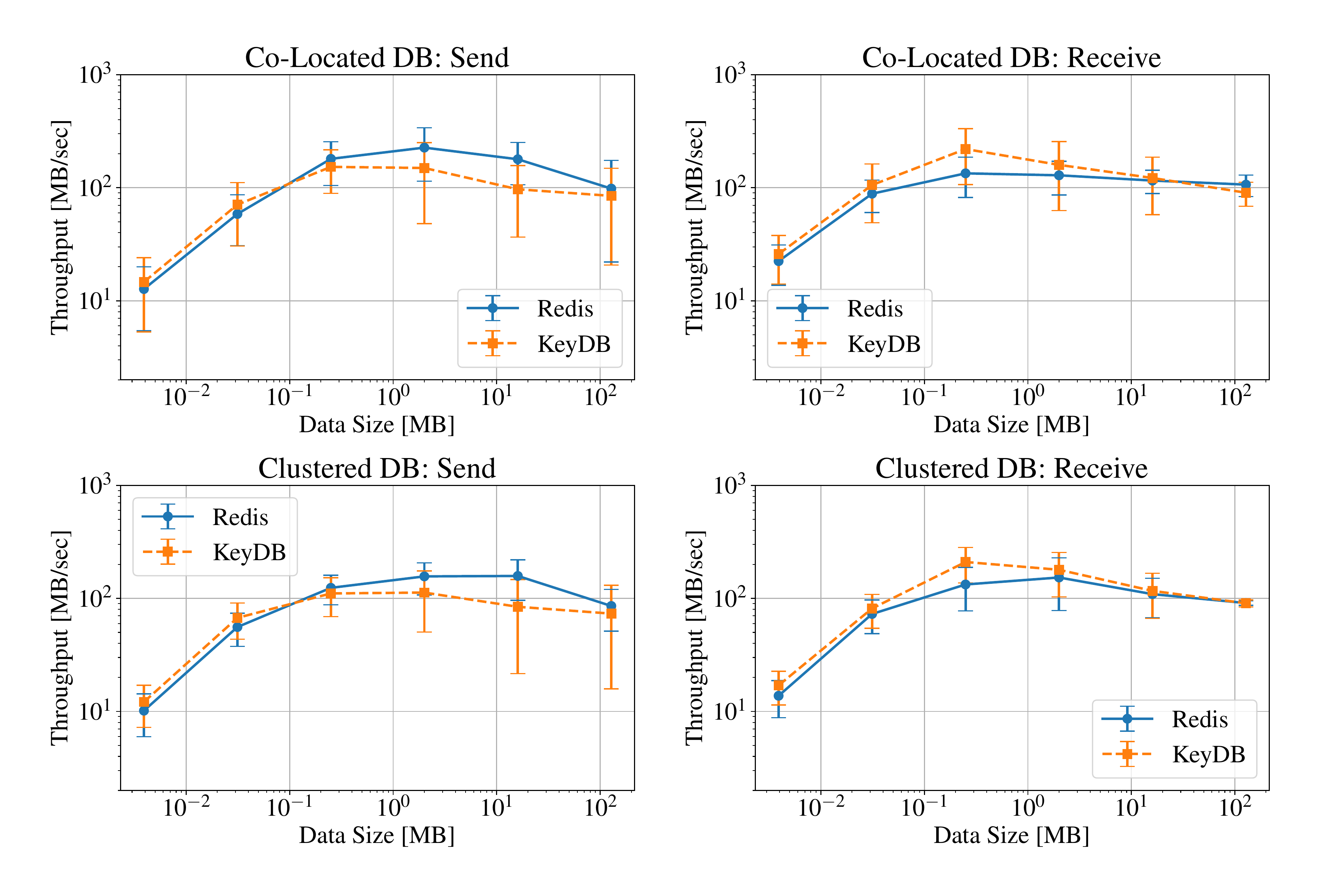}
        \caption{Throughput measurement.}
        \label{fig:data_size_tp}
    \end{subfigure}
    \caption{Cost of data send and retrieve between simulation reproducer with increasing data size using both co-located and clustered deployments.}
    \label{fig:data_size}
\end{figure}

We also investigate how the cost of the data transfer operations is affected by the size of the data being transferred. Both deployments are tested with the Redis and Key DB databases. The simulation reproducer is run in parallel on 24 ranks and data is sent and retrieved for 40 iterations.
The results of this test are presented in Figure~\ref{fig:data_size}, where the cost is expressed in units of time and throughput (i.e., data size per second). 
Note that the term throughput is used loosely to measure the rate at which data is processed by the SmartRedis API. This is because the send and retrieve operations are composed of multiple steps, such as the transfer over the TCP/IP network and writing to memory, and the cost of the individual components was not measured for this paper. 

A few key observations can be made from the results in Figure~\ref{fig:data_size}. First, the performance is very similar between the send and retrieve operations and between the Redis and KeyDB databases. 
Second, in their smallest configurations, the co-located and clustered deployments perform very similarly across the entire range of data sizes tested. This indicates there is no additional penalty incurred by transferring data between nodes over the network at this scale. We conjecture that the similarity in performance is due to the network architecture of Polaris which has a Slingshot 10 interconnect and two 200Gbps network connections per compute node. Consequently, for small applications which do not require scaling to multiple nodes, both deployment strategies are equally performant. 
Third, for data sizes $\ge 256$KB per rank, the cost of the transfer operations increases approximately linearly, resulting in a constant or slightly decreasing throughput. The framework is most efficient for data transfers of size $\ge 256$KB and $\le 16$MB. For small data transfer operations ($< 256$KB), the 
time required to send and retrieve data approaches a constant threshold, resulting in decreasing throughput. We hypothesize that this is because there is a fixed cost to handle an I/O request (regardless of size of the message) within Redis that, for small message sizes, dominates the total cost of the entire operation.
Note that a profiling exercise of the key operations performed by the SmartRedis API to understand the details of this behavior is part of our future work.

\subsubsection{Weak Scaling}
\label{sec:WeakScaling}
After investigating performance of the framework in its smallest configuration, we perform weak scaling tests to understand how the efficiency of the data transfer operations changes as the simulation reproducer scales out.
For these tests, we fix the size of data being transferred on each rank to 256KB and the number of simulation ranks per node to 24. This, therefore, represents the workflow wherein the CFD problem size is increased linearly with the number of ranks used for the simulation. Additionally, we bind the co-located database to 8 CPU cores while the clustered DB is sharded on multiple nodes and allowed to utilize the full socket. 
The cost of the operations is averaged over 40 iterations and over all the simulation ranks, with two extra iterations performed beforehand and discarded as a warmup.
The scale of the runs performed reaches 448 nodes, corresponding to 10,752 simulation ranks, on the Polaris supercomputer with the co-located database and 464 nodes with the clustered database assigning 448 to the simulation and 16 to the sharded database. 

\begin{figure}
    \begin{subfigure}{\columnwidth}
        \centering
        \includegraphics[width=\columnwidth]{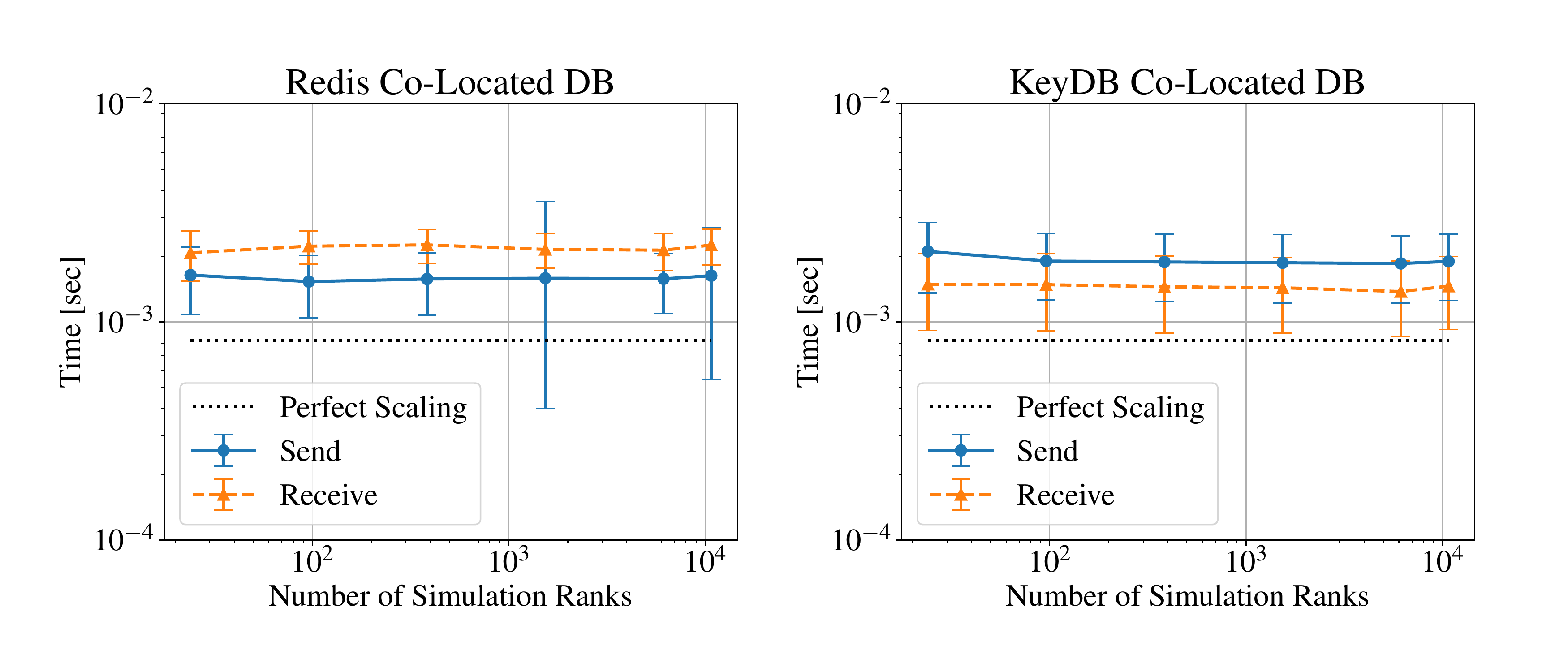}
        \caption{Co-located database.}
        \label{fig:data_weakScale_coDB}
    \end{subfigure}
    \vfill
    \begin{subfigure}{\columnwidth}
        \centering
        \includegraphics[width=\columnwidth]{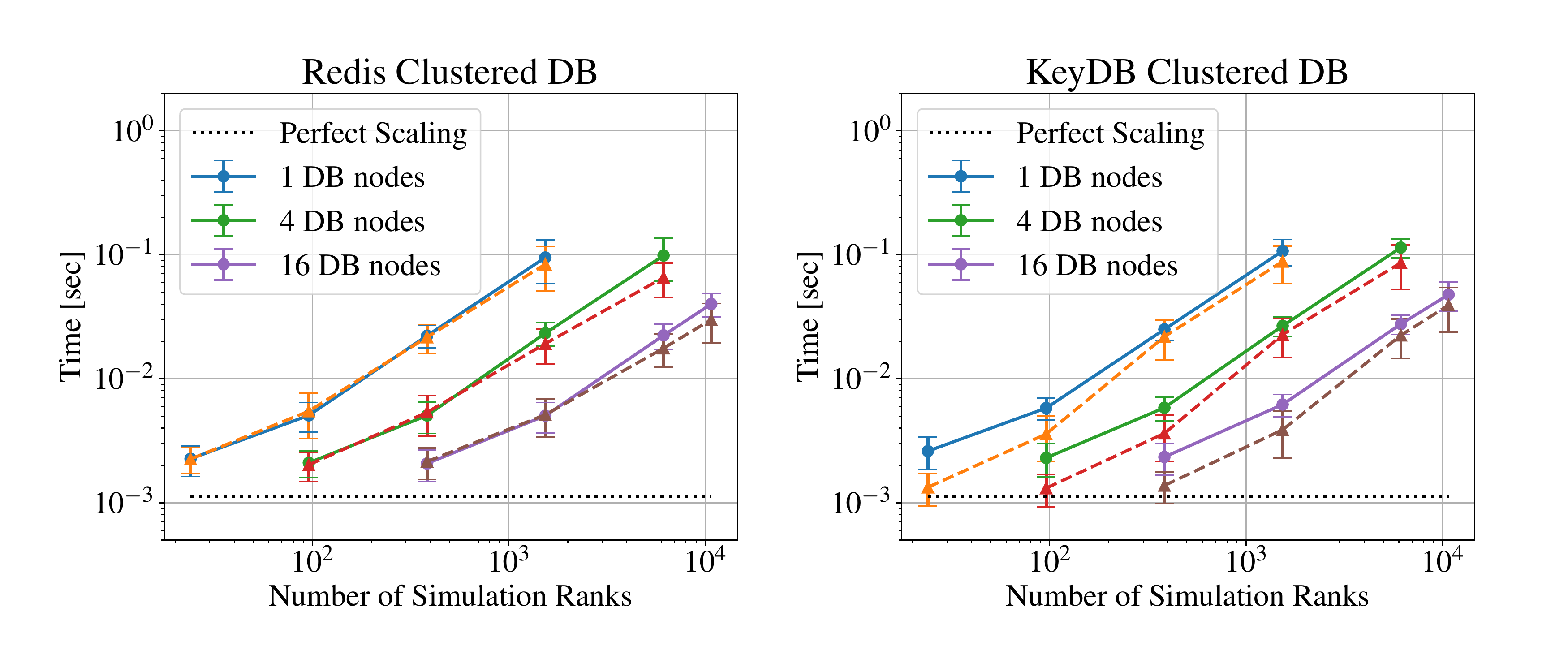}
        \caption{Clustered database. The solid and dashed lines represent send retrieve operations, respectively.}
        \label{fig:data_weakScale_clDB}
    \end{subfigure}
    \caption{Weak scaling of data send and retrieve operations between simulation reproducer database with the co-located and clustered deployments.}
    \label{fig:data_weakScale}
\end{figure}

The results of the scaling tests are shown in Figures~\ref{fig:data_weakScale_coDB} and ~\ref{fig:data_weakScale_clDB} for the co-located and clustered deployments, respectively.
With the former, perfect scaling efficiency is achieved for both the send and retrieve operations and both database software options as demonstrated by the horizontal lines in the figures. This is a key result of our in situ framework and is achieved thanks to the novel deployment of a database co-located on the same nodes as the simulation, thereby eliminating any data transfer between nodes. The overhead of the framework on the simulation is constant and independent of the number of nodes used by the application. As a consequence, our framework does not alter the scaling behavior of the simulation code or distributed training workload it is integrated with, but rather only adds a fixed overhead, which is negligible compared to a flow solver time step or training epoch as demonstrated in Section~\ref{sec:Phasta}.
Moreover, the cost of the send and retrieve operations are very similar to each other showing no bias for one over the other. This means the data producer, which mostly sends data, and the data consumer, which mostly retrieves data, are equally performant when interacting with the database.

By contrast, the weak scaling behavior of the clustered deployment is more complex, as shown in Figure~\ref{fig:data_weakScale_clDB}. 
For a fixed database size, the cost of both data transfer operations increases rapidly as the number of simulation ranks is increased. In fact, the cost appears to increase linearly with the number of ranks after a certain relative threshold (e.g., above 96 ranks with 1 database node). The poor scaling efficiency of this deployment of the framework is due to the bottleneck that occurs at the database when handling the increasing number of client requests (one SmartRedis client is initialized on each simulation rank). Effectively, the database cannot process the requests sufficiently fast and many of the simulation ranks are left idle waiting for the transfer request to complete.

However, this bottleneck can be alleviated by sharding the database to multiple nodes. More specifically, when the size of the database is increased linearly with the number of simulation ranks, the cost of the send and retrieve operations remains approximately constant. This result can be see in  Figure~\ref{fig:data_weakScale_clDB} comparing, for instance, the data transfer time with 1 database node and 24 ranks, 4 database node and 96 ranks, and 16 database nodes and 384 ranks. Moreover, note that the data transfer time is comparable between the two framework deployments strategies when an equal number of nodes is used for the simulation and database in the clustered case, suggesting that transferring data across the network is not a bottleneck for this deployment strategy.
Therefore, when the size of the clustered database is scaled proportionally to the simulation, the framework shows near-perfect scaling efficiency with the this deployment as well. Indeed, this approach is very computationally expensive as it requires twice the number of nodes as the simulation, therefore we recommend the co-located deployment to scale applications.

\subsubsection{Strong Scaling}
\label{sec:StrongScaling}
The scaling efficiency of the data transfer operations with the co-located deployment can also be demonstrated in the context of strong scaling. 
For these tests, the total size of the data being transferred from all simulation ranks to the database is constant at 384MB (approximately the size of the pressure and solution fields of a grid with $230^3$ points), but the on-rank data size is reduced with increasing scale. This, therefore, represents the workflow wherein the CFD problem size is fixed but the scale of the simulation is increased to accelerate the computation. 
The number of simulation ranks per node is fixed to 24 and the co-located database uses 8 CPU cores. Similarly to the weak scaling tests, the scale of the runs performed reaches 448 nodes of Polaris, corresponding to 10,752 simulation ranks.

\begin{figure}
    \centering
    \includegraphics[width=\columnwidth]{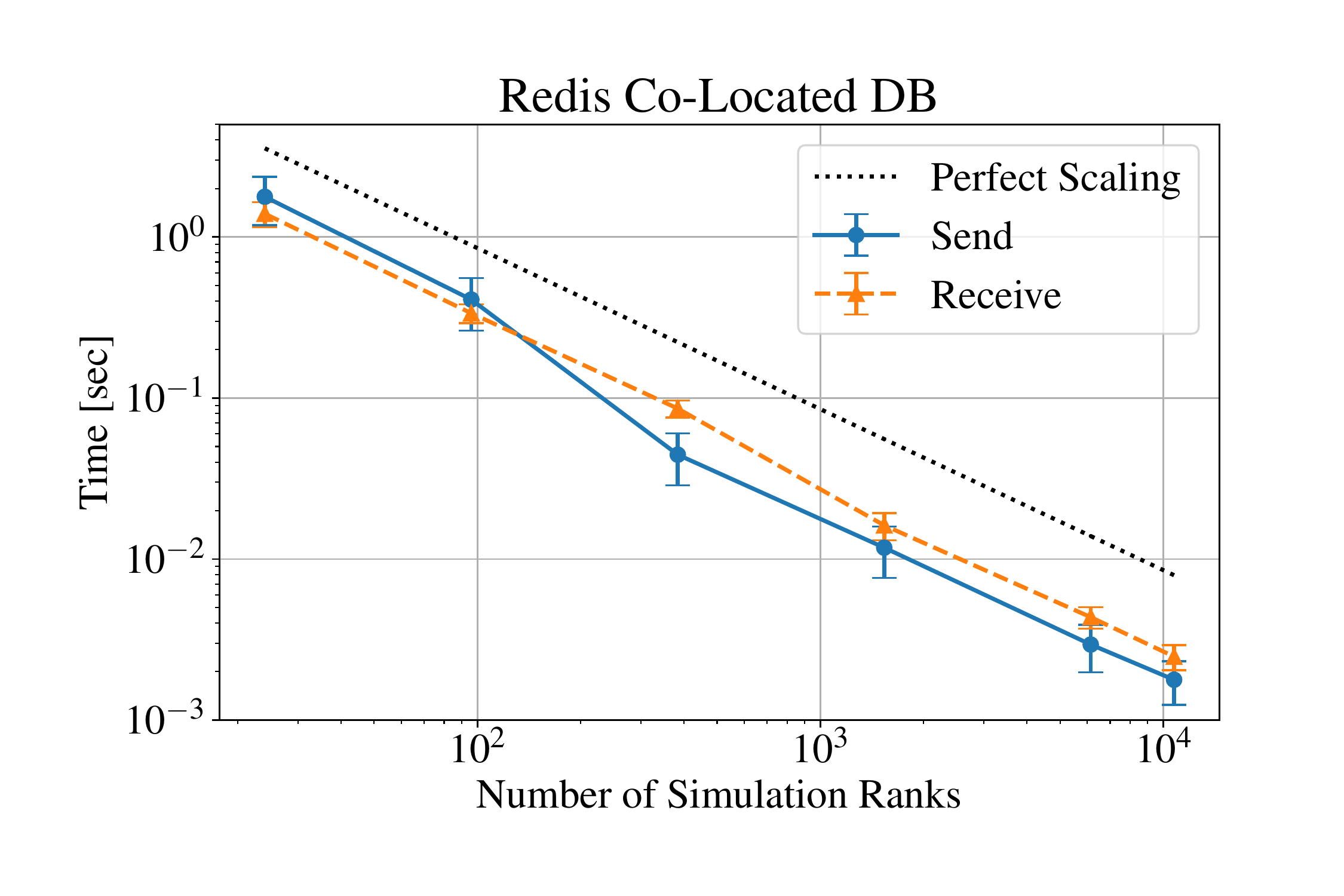}   
    \caption{Strong scaling of data send and retrieve operations between simulation reproducer and the Redis database with the co-located deployment.}
    \label{fig:data_strongScale}
\end{figure}

Figure~\ref{fig:data_strongScale} presents the results of the strong scaling tests with our in situ framework. Only the co-located deployment was used given the superior scalability of this approach and only a Redis database tested given the very similar performance of the two software packages.
Once again, the data transfer operations show perfect scaling efficiency, with the send and retrieve times decreasing linearly with the number of simulation ranks. 
Note, however, that this result is achieved with an on-rank data size $\ge 256$KB. For smaller data sizes, the transfer time approaches a constant value as shown in Figure~\ref{fig:data_size_time}.

%\subsubsection{Effects of Client Request Asynchrony}
%We looked into the effects of asynchrony in the client requests across the simulation ranks. Discuss these here.

\subsection{Model Inference}

In this section, we characterize the performance of evaluating an ML model for inference with our in situ framework. As discussed in Section~\ref{sec:Infra}, inference requires three steps -- sending the inference data, evaluating the model, and retrieving the predicted data. The total cost of performing inference with our framework is therefore the sum of the three components.
Following the discussion above, we selected to use only the Redis database in the co-located deployment for this section. The simulation reproducer and database share the CPU resources in the same way as outline above for the data transfer tests, with the Fortran code using 24 ranks per node and the database being bound to 8 logical CPU cores. 
All 4 GPU on the Polaris nodes are utilized for inference leveraging the RedisAI module and the requests from the simulation ranks are balanced evenly across them by pinning 6 clients to each. The deployment of our framework therefore looks similar to the left panel of Figure~\ref{fig:deployment_inf}.
We use the popular ResNet50 model \cite{He2015} for all inference tests in this section, which requires input data of size $(n,3,224,224)$ and returns predictions of size $(n,1000)$, where $n$ is the batch size.
Results in this section are averaged over 40 iterations and over all the simulation ranks, however two extra iterations are performed beforehand and discarded to warm up the GPU.

\subsubsection{Single Node Performance}

Before investigating the scaling efficiency of our framework, we ground its performance relative to another in situ inference approach on a single node. 
For this purpose, we developed a second reproducer which leverages the C++ LibTorch library. We created C++ functions to load an ML model to the GPU and executing it on inference data to return predictions. Then, using a custom Fortran to C++ bridge, these functions were called from the Fortran main program, thus replicating the equivalent operations provided by the SmartRedis API. This reproducer is an example of a tightly-coupled or \textit{in line} approach and is similar to previous implementations created for CFD solvers \cite{Maulik2020,Maulik2022,Ott2020,Geneva2019}.

\begin{figure}
    \centering
    \includegraphics[width=\columnwidth]{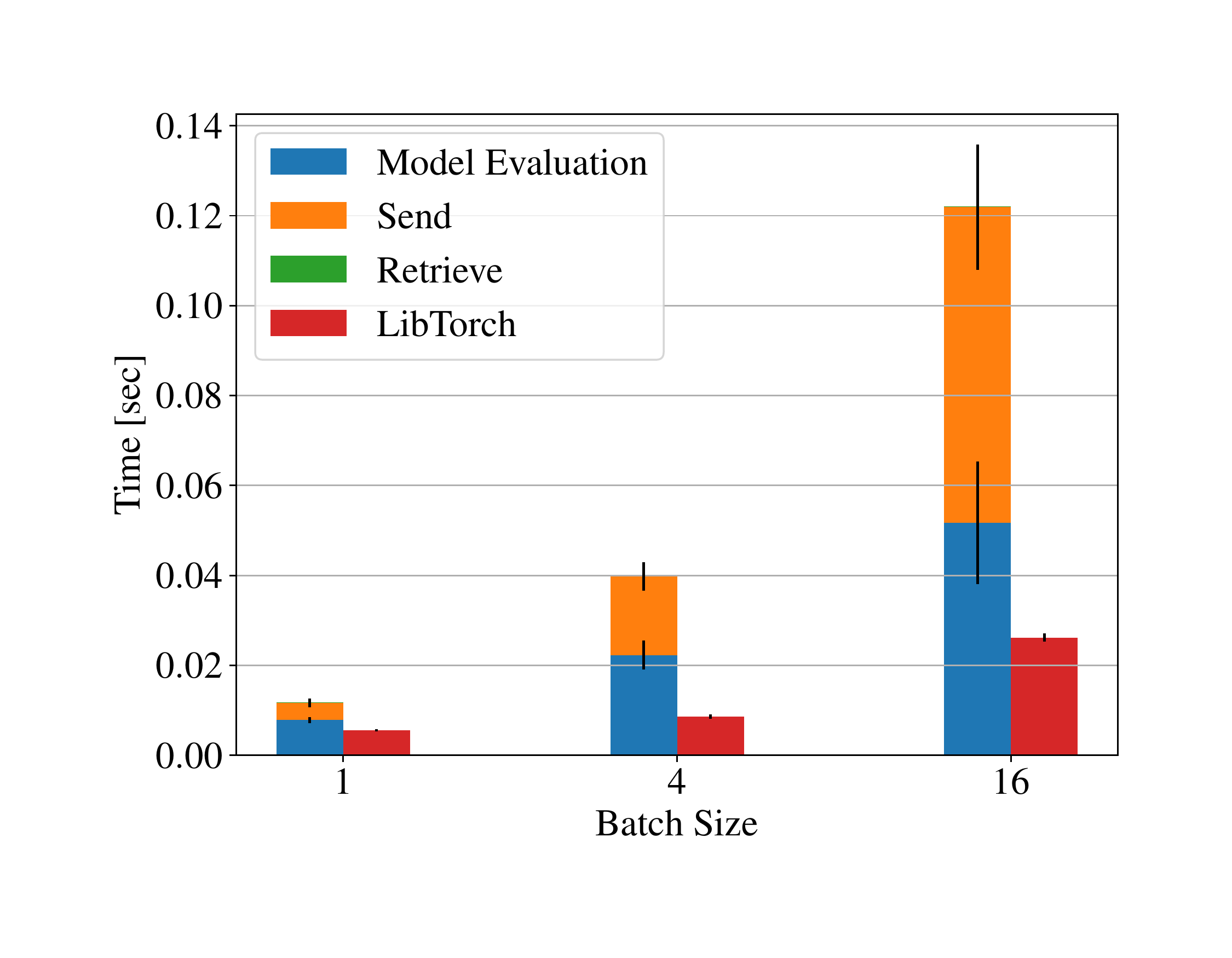}   
    \caption{Cost of the three components required to perform inference with the in situ framework compared to LibTorch.}
    \label{fig:inference_components}
\end{figure}

Figure~\ref{fig:inference_components} compares the performance obtained with our framework to the reproducer using LibTorch for three batch sizes of increasing size. The figure also shows the relative cost of the three operations required by our framework. 
For the ResNet50 model, the send and model evaluation operations are the most expensive of the three components. The smaller size of the array containing the predicted data relative to the input data results in a much faster data transfer. 
%The cost of the send represents an upper bound on the cost of such a request because the reproducer represents a worst case scenario where every client is near simultaneously doing a send request. The retrieval of the inference is negligible can be partially ascribed to staggering of requests that naturally arises due to the asynchronicity that arises as each client completes its send and model evaluation request. 
As expected, the cost of both operations increases with the batch size, however while the data transfer time increases linearly with the size of the data, the model evaluation time does not. As a result, with the larger batch size the data transfer component becomes most expensive, contributing to 57\% of the total cost. 

Comparing the performance of the two reproducers, Figure~\ref{fig:inference_components} indicates that or framework is more costly than the tightly-coupled one using LibTorch. This is both due to the additional cost of data transfer between the simulation and the database and due to the slower model evaluation with RedisAI. With a batch size of 1, using LibTorch offers a speedup of a factor of 2, however this factor increases to 4.6 with batch sizes of 4 and 16. 
While performance is paramount for the efficient deployment of in situ frameworks, we also consider ease of implementation as an important feature to consider. Comparing the Fortran reproducers using our framework to the one using LibTorch under this new metric, our implementation offers a significant advantage. In situ inference capability can be achieved adding less than 10 lines of code with our framework, however creating the Fortran to C++ bridge and creating the the C++ functions calling the LibTorch API required more than 70 lines of code (most of which are devoted to solving problems with C and Fortran interoperability). Moreover, our framework does not limit a simulation to utilize only PyTorch models, whereas the LibTorch implementation does.

\subsubsection{Weak and Strong Scaling}
\begin{figure}
    \centering
    \includegraphics[width=\columnwidth]{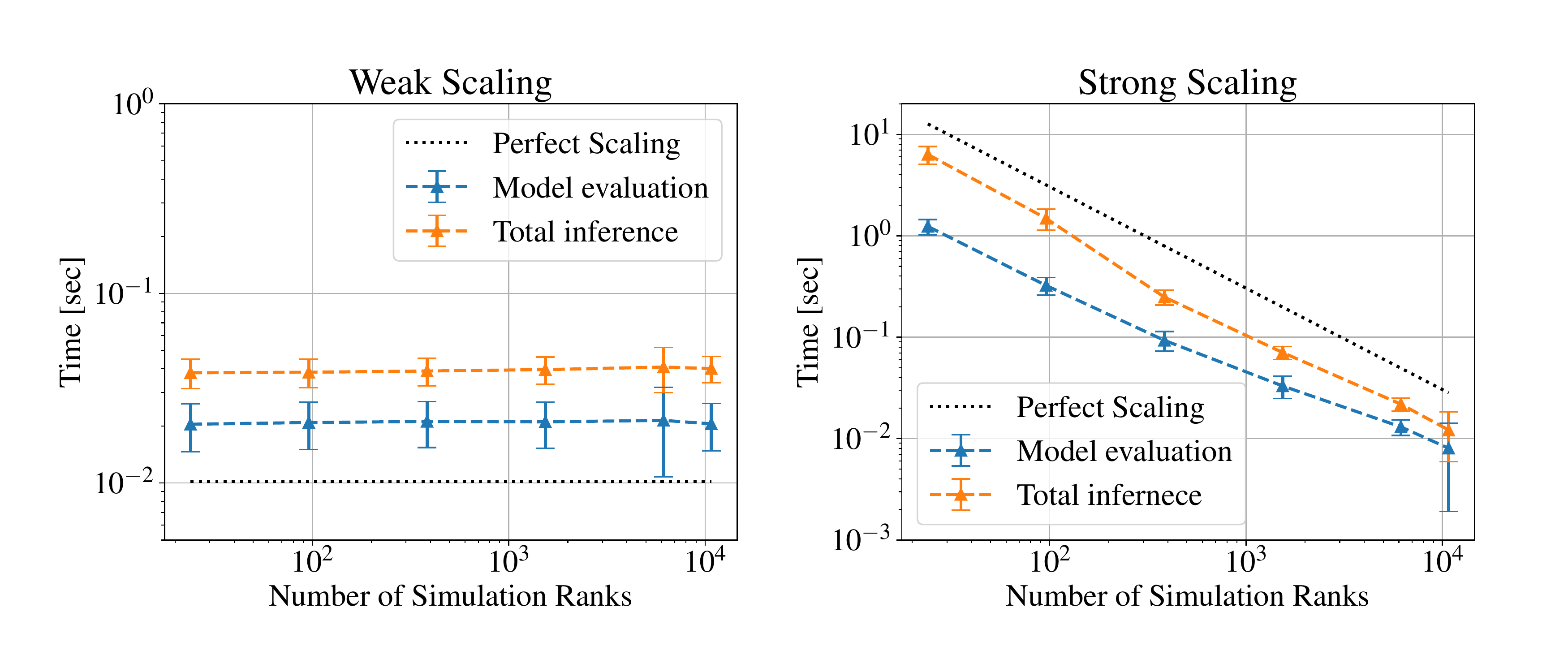}   
    \caption{Weak and strong scaling of in situ inference with the Redis database and co-located deployment.}
    \label{fig:inference_scale}
\end{figure}

We demonstrate the scaling efficiency of our in situ framework during inference tasks by performing similar weak and strong scaling tests as those discussed in Sections~\ref{sec:WeakScaling} and~\ref{sec:StrongScaling}. In the former, the data size and inference workload on each rank are held constant by fixing the batch size, while in the latter the batch size is decreased linearly with the number of ranks. 
Figure~\ref{fig:inference_scale} presents the results obtained for both the model evaluation and total inference costs.
In the case of weak scaling, perfect efficiency is achieved for both quantities. 
In the case of strong scaling, the model evaluation does not achieve a perfect scaling due to the batch size reaching small values at larger scales. However, this performance degradation is amortized by the faster data transfer operations, resulting in a perfect linear scaling behavior of the total inference cost.
This is another key result of this paper and is achieved thanks to our novel embarrassingly parallel deployment of a database co-located on the same nodes as the simulation which eliminates all inter-node data transfer.
Moreover, it demonstrates that our in situ framework achieves perfect scaling efficiency for both training and inference applications.

\section{In Situ Training and Inference of an Autoencoder Model}
\label{sec:Phasta}

After having demonstrated the scalability of our framework and its suitability to tackle large CFD workflows, in this section of the paper we apply our framework to perform in situ training of an autoencoder for compression of turbulent flow states. 
The motivation for this application is to offer a solution to the limitations imposed by the file system storage on HPC clusters, which often prevent researchers from writing and storing as many solution snapshots as required to fully investigate the transient dynamical properties of the flows of interest, either by way of visualization or \textit{post hoc} analysis. With the ability to fine-tune an ML model for a particular class of flow problems, the scientist can train the autoencoder in situ during the initial phase of the simulation and then encode solution snapshots by performing inference during the remainder of the simulation (which in many cases requires a least $\mathcal{O}(10^4)$ time steps), thus storing a much richer time history of the flow solution. 

The CFD solver selected is PHASTA, which is based on a residual-based stabilized, semi-discrete finite element method for the transient,  Navier-Stokes partial differential equations governing fluid flows \cite{Whi03,WhiJan01}. PHASTA is well suited for this work due to its extreme scalability \cite{Rasquin2014} and demonstrated accuracy on a number of DNS \cite{Trofimova2009,Balin2021,Wright2021}. Additionally, it is a CPU only code written mostly in Fortran and thus it is representative of many established CFD solvers. 
The training data consists of instantaneous flow solutions, specifically the pressure and three components of velocity, of a flat plate turbulent boundary layer computed by DNS with an inflow Reynolds number based on the momentum thickness $\theta$ of $Re_\theta=1,000$. The computational domain is discretized into 36 million elements and the simulation setup follows the work of \citet{Wright2021}.

\begin{figure}
    \begin{subfigure}{\columnwidth}
        \centering
        \includegraphics[width=\columnwidth]{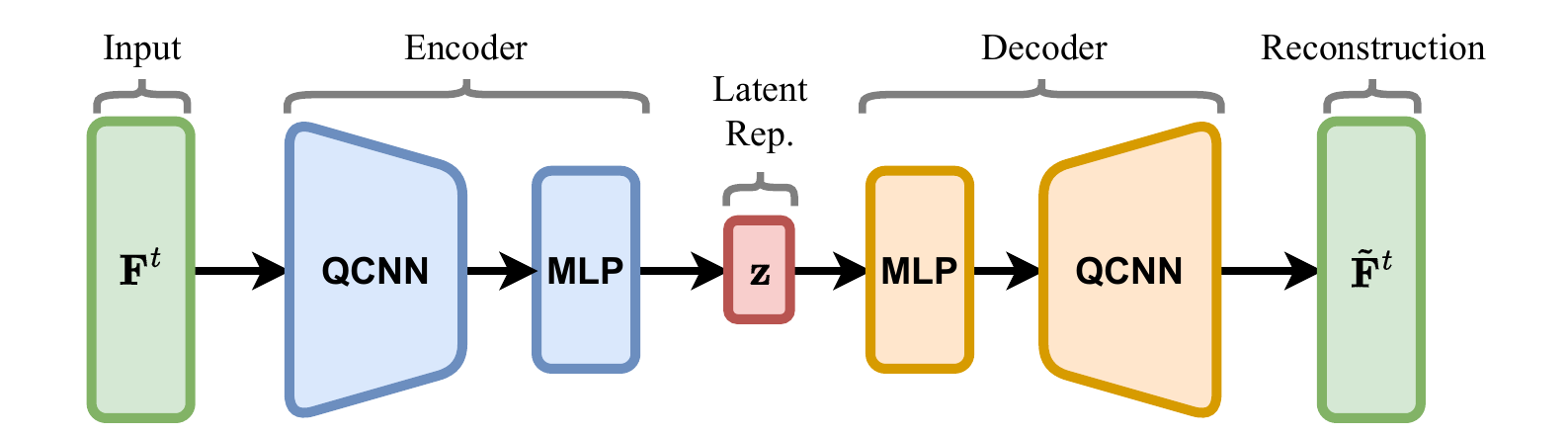}
        \caption{Autoencoder structure. \(F^t\) denotes a single mesh partition input sample from time-step \(t\), and \(\tilde{F}^t\) its subsequent reconstruction.}
        \label{fig:autoencoder}
    \end{subfigure}
    \vfill
    \begin{subfigure}{\columnwidth}
        \centering
        \includegraphics[width=\columnwidth]{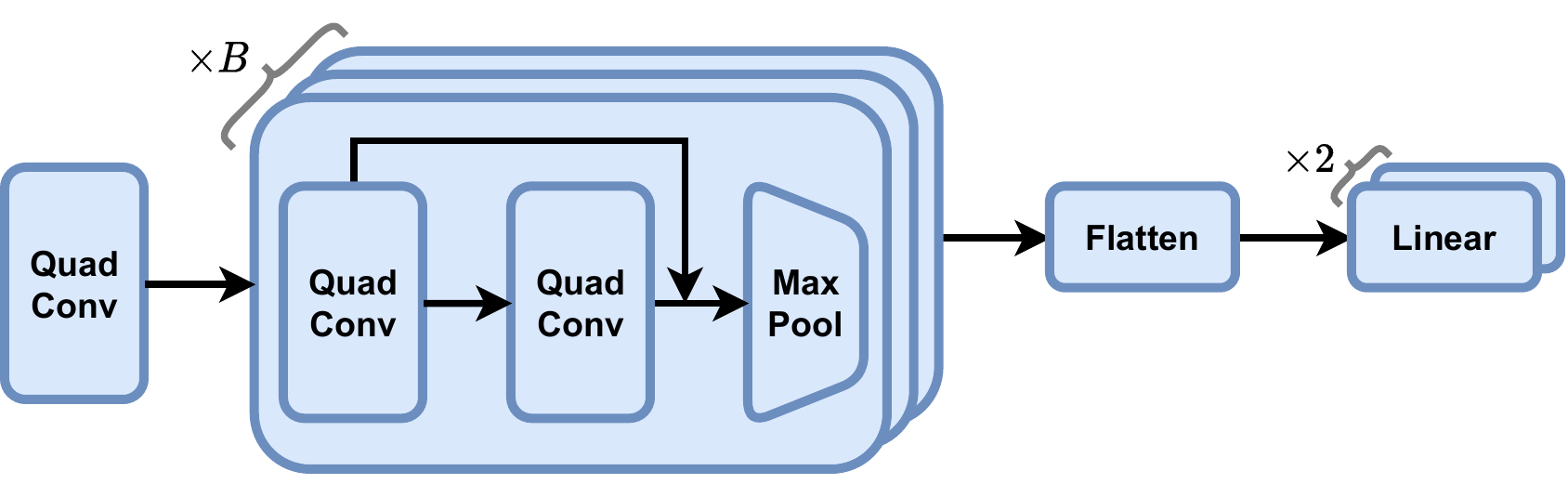}
        \caption{Encoder structure with \(B\) blocks. Activation functions after the QuadConv and linear layers, along with normalization layers, have been excluded for clarity. The linear layers comprise the MLP.}
        \label{fig:encoder}
    \end{subfigure}
    \caption{Model architecture -- figures adapted from \cite{Doherty2023}.}
    \label{fig:model-architecture}
\end{figure}

Our autoencoder is based on the design in \cite{Doherty2023}, and uses the Quadrature-based Convolution (QuadConv) layers presented therein -- made available via the \textit{PyTorch-QuadConv} package \cite{pytorch-quadconv}. The QuadConv operator approximates continuous convolution via quadrature (i.e. a single weighted sum), where both the weights and the kernels are learned during training. The kernels are parameterized by a continuous MLP that maps from spatial coordinates. This framework allows one to apply convolutions directly to unstructured mesh data -- or in this case a non-uniform grid. More details on the QuadConv operator itself can be found in the listed reference. The general structure of our autoencoder remains mostly unchanged, but we have altered various hyper-parameters to suit our needs. The biggest architectural modification is the removal of spectral normalization layers in the MLPs to ensure traceability for online inference. Given the larger scale of the compression task we consider here, we have used two blocks (see Figure~\ref{fig:encoder}) in the encoder and decoder, along with deeper and wider filter MLPs for the QuadConv layers. Specifically, these filters map 3D spatial coordinates through a five layer MLP to \(\mathbb{R}^{16\times16}\) -- where 16 is the number of internal data channels used in the model.
%For complete details on the network configuration, one may consult <cite insitu repo?>. 
Figure~\ref{fig:model-architecture} outlines the overall autoencoder architecture and a more detailed view of the encoder. The decoder is essentially a mirrored version of the encoder, where the flattening and max-pooling layers have been replaced with unflattening and un-pooling respectively. 
Training of the autoencoder is implemented using the PyTorch framework and parallelized with the DistributedDataParallel (DDP) library. The Adam optimizer was chosen for this work with the standard mean squared error (MSE) loss and a learning rate of 0.0001 scaled linearly with the number of ranks. Single precision was used for the data and the model and the latent space dimension was set to 100.

The in situ framework was set up following the results in Section~\ref{sec:Scaling}. A Redis database was used with a co-located deployment, thus following the schematics in Figure~\ref{fig:deployment}. During training, the simulation was run in parallel with 960 ranks and the distributed training workload used 160 GPU, with one rank per GPU. On each of the 40 Polaris nodes, the simulation ran on the CPU using 24 cores, the database was pinned to 8 cores, and the model training utilized the 4 available GPU, thus taking advantage of all the compute resources available. 
%During inference, the 4 GPU on each node were used to evaluate the encoder model leveraging the SmartRedis API and the RedisAI module.

Training was carried out for 500 epochs, during which the overhead of the in situ infrastructure on both the PHASTA simulation and distributed training workload was measured. These results are summarized in Tables~\ref{tab:phasta} and~\ref{tab:train}, where we present the average and standard deviation across all ranks of the total time spent in each component.
On the simulation side, each PHASTA rank sends the instantaneous pressure and velocity fields of its owned domain partition to the co-located database. This data transfer involves sending 690KB of data and is performed every two time steps. Nevertheless, the overhead of this operation is negligible compared to the cost of advancing the solution (the sum of equation formation and solution in Table~\ref{tab:phasta}). In fact, the total overhead of the in situ framework, composed of the client initialization and all data transfers, is $\ll 1\%$ of the PDE integration time. Note that while the training data was transferred every two time steps in this case, the cost would remain negligible even if this transfer was made after every step.

On the distributed training side, each rank gathers the training data produced by PHASTA from the co-located database at the beginning of each epoch. Due to the ratio of 24 simulation ranks to 4 ML ranks per node, 6 arrays of training data are gathered and concatenated before the distributed mini-batch stochastic gradient descent optimization is applied to converge the model weights. As shown in Table~\ref{tab:train}, the overhead of this data transfer operation is just over 1\% of the total training time, and thus can be considered negligible. The cost of transferring metadata to and from the database is larger in this case (4.4\% of the training time) because the ML workload must query the database multiple times while waiting for the first training snapshot to be provided by PHASTA. 

\begin{table}
  \caption{PHASTA solver components during in situ training avraged across ranks}
  \label{tab:phasta}
  \begin{tabular}{ccl}
    \toprule
    Solver Component & Average [sec] & Standard Deviation [sec] \\
    \midrule
    Equation formation & 45.426 & 0.678 \\
    Equation solution & 453.386 & 0.698 \\
    Client initialization & 0.002 & 0.001 \\
    Metadata transfer & 0.065 & 0.005 \\
    Training data send & 0.120 & 0.021 \\
  \bottomrule
\end{tabular}
\end{table}

\begin{table}
  \caption{ML training components during in situ training averaged across ranks}
  \label{tab:train}
  \begin{tabular}{ccl}
    \toprule
    Training Component & Average [sec] & Standard Deviation [sec] \\
    \midrule
    Total training & 332.700 & 0.040 \\
    %Batch computation & 277.238 & 0.336 \\
    Client initialization & 0.002 & 0.001 \\
    Metadata transfer & 14.789 & 0.093 \\
    Training data retrieve & 4.455 & 0.301 \\
  \bottomrule
\end{tabular}
\end{table}

\begin{figure}
    \centering
    \includegraphics[width=\columnwidth]{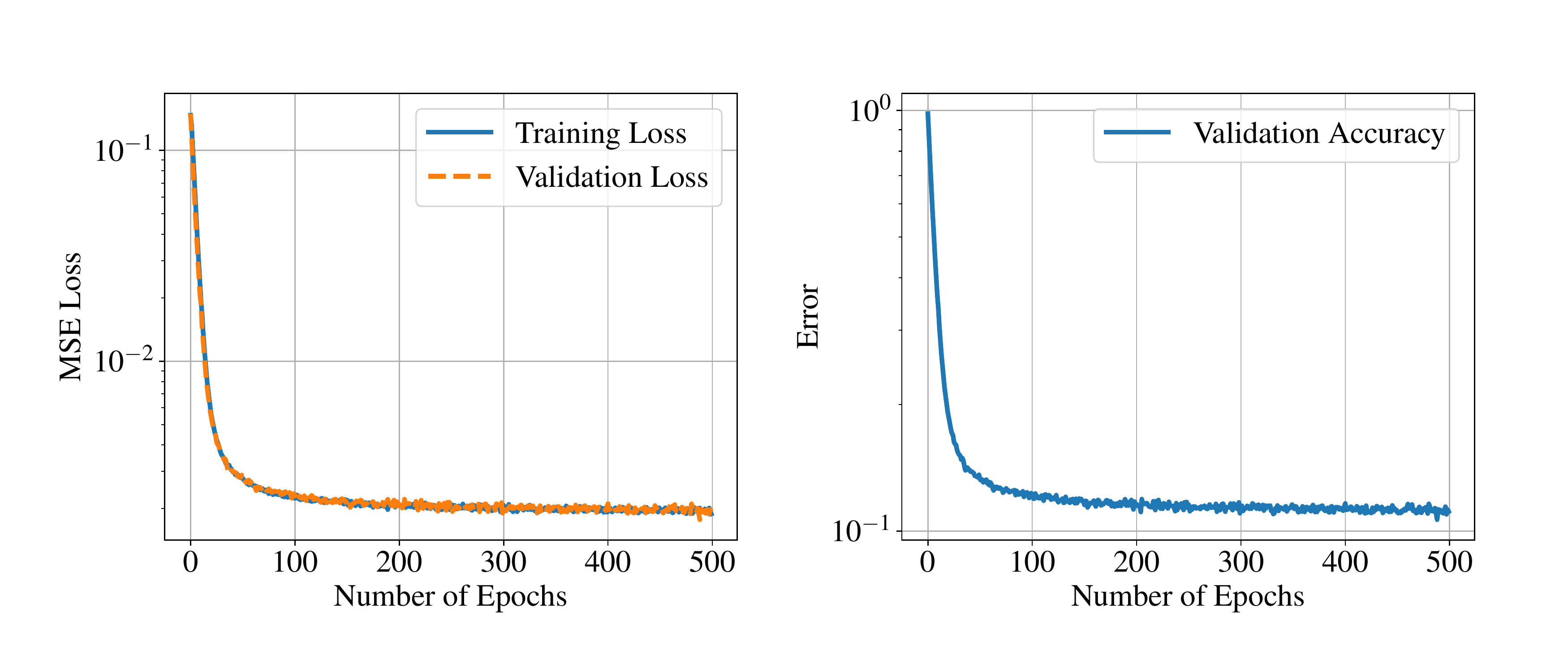}   
    \caption{Convergence of the training loss, validation loss and validation error during in situ training of the QuadConv autoencoder model.}
    \label{fig:training_loss}
\end{figure}

The outcome of in situ training of the autoencoder is shown in Figure~\ref{fig:training_loss}, which reports the training and validation losses as well as the validation relative reconstruction error defined as the relative Frobenius norm
\begin{equation}
  \frac{1}{T}\sum_{t=1}^T\frac{\sqrt{\sum_{ij}(F^t_{ij} - \tilde{F}^t_{ij})^2}}{\sqrt{\sum_{ij}(F^t_{ij})^2}} .
  \label{eq:error}
\end{equation}
Note that in Equation~\ref{eq:error}, as in Figure~\ref{fig:autoencoder}, \(F^t\) and $\tilde{F}^t$ are the input data and its reconstruction for samples \(t=1,\ldots,T\). The data \(F^t\in\mathbb{R}^{C\times N}\) has \(C\) (in our case 4) channels and \(N\) points, indexed by \(i\) and \(j\) respectively. Validation is performed on one of the six tensors grabbed by each ML rank at random from the database at the beginning of each epoch. 
As seen in the figure, the training and validation loss converge smoothly and decrease by two orders of magnitude. This was achieved with the training completing approximately 20 epochs on each solution snapshot supplied by PHASTA, however the model convergence was not observed to be sensitive to this synchronization parameter.
The validation error also converges smoothly, but in this case decreasing by a single order of magnitude to 10\%. After training, we continue to integrate the simulation to produce testing samples for the model. The testing error remains consistent with the final validation error at 10\%. 

The magnitude of the reconstruction error obtained warrants further discussion. We note it is larger than what was reported by \citet{Doherty2023}, however in their work the autoencoder was applied to two-dimensional data of lower complexity in an offline setting. The nature of the turbulence present in this flow introduces three-dimensionality, a wide range of spatial and temporal scales, and large gradients in the flow variables. 
Moreover, the dimension of the latent space was set to 100, resulting in a $1700\times$ spatial compression of the simulation data. The compression factor is obtained by dividing the size of the simulation data on each PHASTA rank by the latent space dimension. This value is extreme, and it is likely better performance could be obtained by decreasing the compression factor via a larger latent dimension. In our experiments, we tried a latent dimension of 500 (\(340\times\) compression), but were unable to obtain meaningful improvements in reconstruction error. It is unclear at this time what the limiting factor is and  part of our future work will focus on investigating how to increase the model performance on turbulent flow data of this kind. 
Nevertheless, we note that this compression task was designed to validate our in situ framework by applying it to a realistic CFD problem and demonstrate its negligible overhead on the simulation and training code. The results presented in this section support the completion of those goals.

% Paragraph on inference on same problem but future time steps, mentioning accuracy of reconstruction and showing reconstruction.

\section{Conclusions and Future Work}
\label{sec:Conclusion}

In this paper, we presented a framework that facilitates the coupling of CFD simulations with machine learning and enables in situ training and inference of ML models at scale. The framework leverages the SmartSim library to deploy a database which stores inference and training data, useful metadata, and ML models in memory for the duration of the run, thus circumventing the file system. 
Additionally, the database allows the simulation and distributed training to communicate asynchronously, leading to a loosely-coupled implementation that minimizes idle time during training. Lastly, the database utilizes the RedisAI module to load and evaluate ML models on the GPU to obtain predictions during inference.

Thanks to a novel deployment of the database which co-locates it on the same nodes as the simulation and training applications, our proposed framework eliminates all data transfer between nodes and makes full use of the compute resources on heterogeneous systems. On the Polaris supercomputer located at the Argonne Leadership Computing Facility, our framework obtained perfect weak and strong scaling efficiency of the data transfer and inference costs up to the full size of the machine. Therefore, our framework does not alter the scaling behavior of the simulation code or the distributed training workload it is integrated with, but
rather only adds a fixed overhead regardless of the number of nodes used for the application. This overhead was demonstrated to be negligible relative to the cost of integrating a solution time step or computing a training epoch by applying our framework to train an autoencoder in situ from turbulent flat plate data simulated by DNS.

Future work will be focused towards improving the reconstruction accuracy of the autoencoder model for turbulent flow data and extending it to support distributed training from unstructured grids. Additionally, we will explore using transfer learning to fine-tune an offline, pre-trained model to a variety of flow problems with the goal of accelerating the in situ learning process. Finally, our in situ framework will be applied to train and deploy turbulence closure models for large eddy simulation of wall-bounded turbulent flows.

% ongoing and future work: 1) improve the accuracy of the encoder model on turbulent flows, 2) call the model from PHASTA to save snapshots, 3) explore transfer learning ideas where a base model is trained offloine on small canonical dataset and then fine tuned on a series of turbulence flow problems.

%\begin{figure}[h]
%  \centering
%  \includegraphics[width=\linewidth]{sample-franklin}
%  \caption{1907 Franklin Model D roadster. Photograph by Harris \&
%    Ewing, Inc. [Public domain], via Wikimedia
%    Commons. (\url{https://goo.gl/VLCRBB}).}
%  \Description{A woman and a girl in white dresses sit in an open car.}
%\end{figure}

\begin{acks}
This research used resources of the Argonne Leadership Computing Facility, which is a DOE Office of Science User Facility supported under Contract DE-AC02-06CH11357. Additionally, it was supported by the Office of Science, U.S. Department of Energy, under Contract DE-AC02-06CH11357 and is associated with an ALCF Aurora Early Science Program project.
\end{acks}

\bibliographystyle{ACM-Reference-Format}
\bibliography{biblio}

%\appendix

\end{document}